\documentclass[runningheads]{llncs}

\usepackage{eccv}

\usepackage{eccvabbrv}

\usepackage{graphicx}
\usepackage{booktabs}
\usepackage{multirow}
\usepackage{wrapfig}

\usepackage[accsupp]{axessibility}  

\usepackage{hyperref}

\usepackage{orcidlink}

\begin{document}

\title{When Token Compression Breaks: Structural Pruning vs. Token Reduction for Robust ViT Segmentation under High Compression} 

\titlerunning{Token Compression vs. Structural Pruning for ViT Segmentation}

\author{Tien-Phat Nguyen
\orcidlink{0000-0002-4204-842X} \and
Ngai-Man Cheung
\orcidlink{0000-0003-0135-3791}}

\authorrunning{T.-P.~Nguyen and N.-M.~Cheung}

\institute{Temasek Laboratories, Singapore University of Technology and Design, Singapore \\
\email{\{tienphat\_nguyen, ngaiman\_cheung\}@sutd.edu.sg}}

\maketitle

\begin{abstract}

Vision Transformers (ViTs) are strong backbones for semantic segmentation, but their computational cost limits deployment. 
Recent token compression methods for efficient transformer-based segmentation reduce this cost by decreasing the number of tokens. However, existing evaluations primarily focus on low-to-moderate compression, leaving their behavior under aggressive compression and corrupted inputs unclear.
Meanwhile, structural pruning provides an orthogonal route to efficiency by removing redundant components in the ViT architecture, but is rarely compared to token compression under a unified protocol.
To bridge this gap, we benchmark representative token compression and structural pruning methods for ViT-based semantic segmentation under matched FLOPs on ADE20K and Cityscapes, together with their common-corruption variants ADE20K-C and Cityscapes-C. 
Our results reveal a consistent trend on both clean and corrupted inputs: token compression is highly effective at mild reductions but degrades sharply when compression becomes severe, consistent with substantial information loss from overly aggressive token reduction. In contrast, structural pruning exhibits a smoother degradation curve and is more stable at high compression.
Motivated by these findings, we study a \emph{prune-then-merge} pipeline that applies moderate token compression on top of a moderately pruned backbone. At comparable FLOPs, this combined strategy consistently achieves a better accuracy-robustness trade-off at high compression, offering a practical recipe for deployment-oriented ViT segmentation.
Code is available at \url{https://github.com/phatnguyencs/vit-seg-compression}.
 
\end{abstract}

\section{Introduction}
\label{sec:intro}

\begin{figure}[t]
  \centering
  \includegraphics[width=0.6\textwidth]{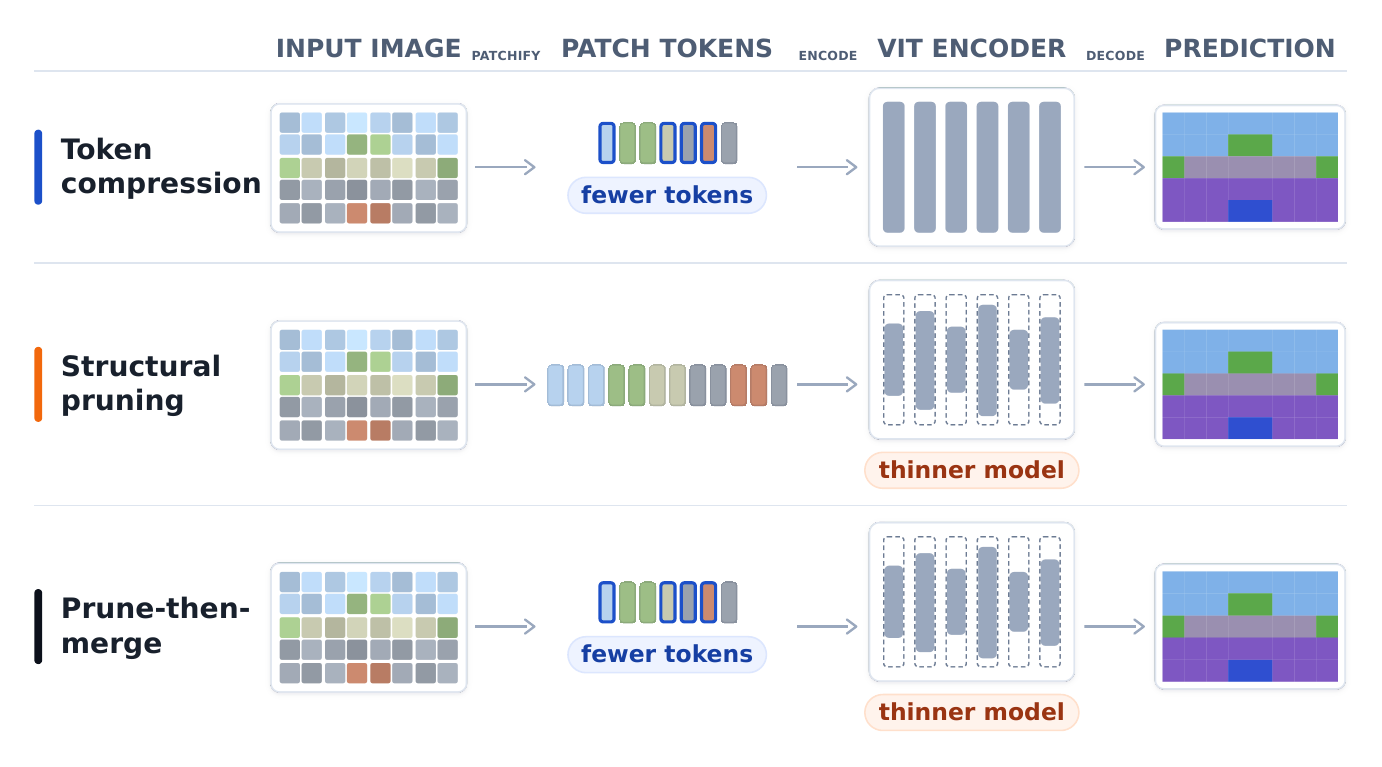}\hfill
  \includegraphics[width=0.4\textwidth]{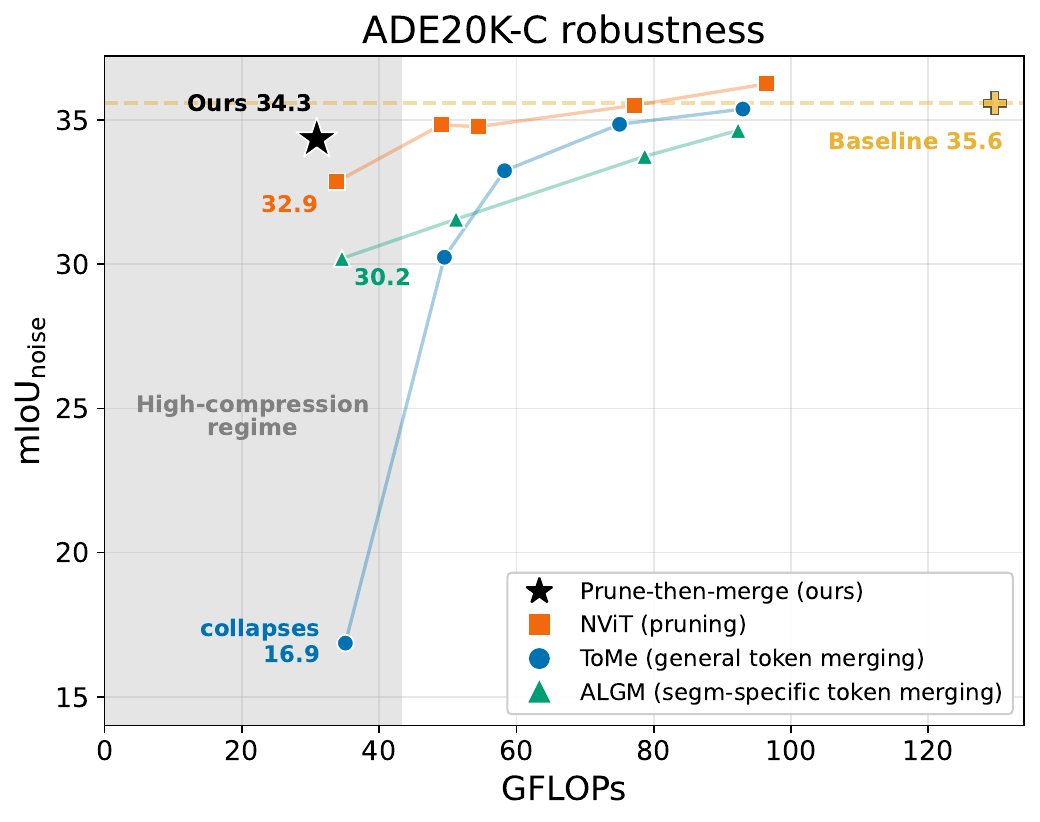}
  \caption{
    \textbf{Compression strategies and their robustness-compute trade-off.}
    \emph{Left:} token compression reduces the number of processed tokens, structural pruning reduces model width/capacity, and our \emph{prune-then-merge} pipeline combines both techniques.
    \emph{Right:} a summary view of the robustness-compute trade-off on ADE20K-C, measured by corrupted-input mIoU versus compute.
    Token compression can preserve performance at mild reductions but drops sharply in the high-compression regime.
    Structural pruning degrades more smoothly, while \emph{prune-then-merge} provides a stronger trade-off at comparable FLOPs.
  }

  \label{fig:intro}
\end{figure}

Vision Transformer (ViT)~\cite{vits} encoders are now a common choice for semantic segmentation, producing high-performance architectures \cite{segmenter,SETR,segformer,sam}. However, their deployment remains challenging as self-attention scales quadratically with token count, and high-resolution segmentation requires processing large spatial token grids~\cite{efficient_vit}. 
This motivates compression and acceleration techniques that reduce computation while preserving segmentation quality.
At the same time, real-world segmentation systems must remain reliable under distribution shifts such as noise, blur, and weather artifacts \cite{data_c}.
Therefore, clean accuracy and FLOPs alone are insufficient for evaluating compressed segmentation models, and corruption robustness must also be considered. 

To improve the efficiency of ViT segmentation models, two major compression strategies have been widely explored.
Token compression~\cite{algm,cts,dtop,dovit} is a data-adaptive strategy that reduces the number of tokens processed by the transformer, for example by pruning redundant tokens or merging spatially similar tokens.
Structural pruning~\cite{nvit,upop,isomorphic}, in contrast, is an architecture-centric strategy that removes redundant model components, such as attention heads, embedding dimensions, or FFN channels, to obtain a permanently compact backbone.
\cref{fig:intro} left summarizes these two strategies and our \emph{prune-then-merge} combination, which applies token compression on top of a structurally pruned backbone.

Despite extensive progress in both compression directions, two questions remain underexplored for semantic segmentation.
First, representative token compression and structural pruning methods are rarely compared under a unified segmentation pipeline with matched FLOPs. 
This makes it difficult to assess their performance and corruption robustness across a wide range of compression levels.
Because these methods act on different dimensions (tokens versus network capacity), their degradation behaviors can differ substantially, especially under aggressive compression.
Second, although pruning and token compression are potentially complementary, it remains unclear whether applying moderate token compression on top of a pruned backbone yields a stronger accuracy-robustness trade-off than using either technique alone at comparable FLOPs.

To address these gaps, we benchmark representative token compression and structural pruning methods within a controlled flat-token ViT segmentation pipeline on ADE20K~\cite{ade20k} and Cityscapes~\cite{cityscapes}, together with their corruption variants ADE20K-C and Cityscapes-C~\cite{data_c}, under matched-FLOPs comparisons across a wide range of compression levels.
Our benchmark shows that the two compression strategies exhibit distinct degradation behaviors.
Token compression can preserve performance at low-to-moderate reduction, particularly for segmentation-aware methods, but degrades sharply under severe compression, with substantial drops in both clean performance and corruption robustness.
In contrast, structural pruning exhibits a smoother degradation curve and is generally more stable at high compression.
To better understand this behavior, we further analyze feature diversity and qualitative predictions, showing that aggressive token compression can reduce representation rank and produce less coherent predictions around fine structures and local boundaries.
Motivated by these findings, we study a \emph{prune-then-merge} pipeline: first apply moderate structural pruning to obtain a compact backbone, then apply moderate token compression on top.
Under comparable FLOPs, this stacked strategy yields a stronger accuracy-robustness trade-off at high compression, providing a practical recipe for ViT segmentation.
\cref{fig:intro} right provides an overview of these findings from the robustness-compute perspective on ADE20K-C.

In summary, our contributions are as follows:
\begin{itemize}
    \item \textbf{Unified benchmark across compression regimes.} We benchmark representative token compression and structural pruning methods for flat-token ViT segmentation under matched FLOPs on ADE20K, Cityscapes, and their corrupted variants. Our results show that token compression can preserve performance at low-to-moderate reduction but becomes fragile at high compression, while structural pruning is more stable and better preserves corruption robustness.
    We complement accuracy-compute curves with effective-rank analysis and qualitative examples, showing that aggressive token compression can reduce feature diversity and produce less coherent predictions around fine structures and local boundaries.
    \item \textbf{Practical prune-then-merge strategy.} We study a stacked pipeline that applies moderate token compression on top of a pruned backbone, achieving a stronger trade-off between accuracy and robustness than using either strategy alone under comparable compute at high compression.
\end{itemize}

\section{Related Work}

\subsection{Vision Transformers for Semantic Segmentation}
Vision Transformers were first introduced to semantic segmentation by SETR~\cite{SETR}, which repurposed a standard ViT encoder for dense prediction by formulating segmentation as a sequence-to-sequence task. Segmenter~\cite{segmenter} further established ViTs as effective segmentation backbones by adopting a fully transformer-based architecture with a mask transformer decoder, achieving strong results on ADE20K and Cityscapes while explicitly highlighting the computational cost of global self-attention at high resolution. SegFormer~\cite{segformer} addressed part of this issue by introducing a hierarchical transformer encoder and a lightweight decoder, improving efficiency while maintaining competitive accuracy. Subsequent segmentation frameworks~\cite{maskformer,mask2former,oneformer} have further advanced segmentation quality through improved decoder designs, training formulations, and multi-task unification, while typically relying on strong pretrained transformer backbones.

An alternative direction focuses on designing efficient transformer architectures from scratch. TopFormer~\cite{topformer} targets mobile settings through token pyramids, and EfficientViT~\cite{efficient_vit} replaces softmax attention with linear attention to significantly reduce latency. While effective, these approaches require architectural redesign and retraining. 
Compression of pretrained backbones provides a complementary route for efficient segmentation, since it can reuse pretrained ViT backbones and adapt them to lower compute budgets rather than requiring a newly designed architecture~\cite{survey}.
Our work follows this direction by compressing pretrained ViT backbones within a controlled flat-token segmentation pipeline.

\subsection{Token Compression for Efficient Vision Transformers}
Token-level compression lowers the cost of transformer inference by reducing the number of tokens being processed, either through token pruning or token merging.
Many approaches are developed and validated primarily on image classification, and some are explicitly designed for classification. EViT~\cite{evit} uses CLS-to-token attention for selection, while ATS~\cite{ats} and SPViT~\cite{spvit} learn recognition-oriented token selection policies. In contrast, several methods define compression using more task-independent criteria that do not rely on a classification head, such as ToMe~\cite{tome} (feature-space similarity merging), Token Fusion (ToFu)~\cite{tofu} (bridging pruning and merging), and DiffRate~\cite{DiffRate} (learning adaptive layer-wise reduction rates).
However, token compression strategies developed for classification do not transfer straightforwardly to semantic segmentation, where dense prediction requires preserving spatial coverage for background regions and boundaries~\cite{cts,algm,dtop}.
To address this mismatch, segmentation-aware methods design criteria tailored to dense prediction: CTS~\cite{cts} merges locally redundant tokens guided by semantic consistency, ALGM~\cite{algm} performs local-to-global merging with adaptive token budgets across layers, and DToP~\cite{dtop} reduces computation via early exiting of confidently predicted tokens.

Despite these advances, evaluations remain fragmented and often focus on low-to-moderate compression ratios, where many methods preserve accuracy. Recent work~\cite{token_benchmark} provides a comparative study of token reduction but focuses on classification benchmarks and does not consider segmentation. 
For semantic segmentation, token compression under high compression remains less explored, although severe token reduction may cause substantial information loss and degrade robustness.
Moreover, prior work rarely compares token compression against structural pruning under matched FLOPs within the same segmentation pipeline, especially under common corruptions.

\subsection{Structural Pruning for Vision Transformers}
Structural pruning reduces inference cost by reducing architectural components (e.g., attention heads, embedding/MLP channels, or layers), producing compact models with fewer FLOPs and parameters. Recent methods adopt principled importance criteria for cross-component pruning, such as NViT~\cite{nvit}, which uses Hessian-based saliency with latency-aware guidance, and SAViT~\cite{savit}, which models joint importance across heterogeneous structures. 
Beyond ViT-specific designs, architecture-agnostic pruning methods improve portability across model families.
DepGraph~\cite{depgraph} achieves this by automatically modeling layer/channel dependencies during pruning.
Isomorphic Pruning~\cite{isomorphic} makes importance scores comparable across different substructures, allowing a unified pruning criterion that can be used across architectures such as ViTs~\cite{vits} and ConvNeXts~\cite{convnext}.

For semantic segmentation, structural pruning has shown promising results. UPop~\cite{upop} reports pruning on Segmenter for ADE20K, and NViT~\cite{nvit} demonstrates that ImageNet-pruned ViTs retain competitive segmentation performance after fine-tuning. 
However, these studies mainly treat pruning as a standalone compression strategy, without systematically comparing it with token compression or evaluating its robustness under common corruptions.

The efficiency-robustness trade-off has been studied in the classification setting. Prior work~\cite{compressed_forget} shows that compressed networks can retain similar clean accuracy while disproportionately degrading on atypical, noisy, and long-tail examples, and that high compression amplifies sensitivity to distribution shifts such as ImageNet-A and ImageNet-C. 
We extend this perspective to ViT-based semantic segmentation by evaluating two orthogonal compression axes, token reduction and structural pruning, with explicit corruption-robustness evaluation.

A few works~\cite{vtclfc,multi_dim_pruning} explore multi-axis compression by jointly pruning architectural and token dimensions with joint objectives.
In contrast, we study a practical pipeline \emph{prune-then-merge} and evaluate this strategy under matched compute in the high-compression regime.

\section{Benchmark Protocol}
\label{sec:benchmark}
This section describes our unified benchmark for comparing two orthogonal efficiency strategies in ViT-based semantic segmentation: token compression and structural pruning. 
We first state the benchmark objectives and research questions (\cref{sec:benchmark_objectives}). 
We then introduce the controlled experimental setup, including the segmentation pipeline, evaluated methods, and training protocol, and efficiency/evaluation metrics (\cref{sec:benchmark_setup}). 
Finally, we describe the clean datasets, corrupted variants, and how corrupted-input robustness is measured (\cref{sec:benchmark_dataset}).

\subsection{Benchmark objectives}
\label{sec:benchmark_objectives}
We benchmark token compression and structural pruning within ViT-based semantic segmentation, and evaluate both clean accuracy and robustness under common corruptions.
Beyond measuring accuracy under compute constraints, our goal is to characterize how robustness evolves as compression becomes more aggressive under matched FLOPs.
Concretely, we ask:
\textbf{(Q1)} How do token compression and structural pruning trade off clean accuracy and corruption robustness across compute budgets?
\textbf{(Q2)} Which strategy degrades more gracefully in the high-compression regime?

\subsection{Experiment setup and evaluation protocol}
\label{sec:benchmark_setup}
\paragraph{\textbf{Segmentation pipeline.}} 
All methods are evaluated under the same segmentation pipeline for controlled comparison. 
We adopt a ViT-based segmenter \cite{segmenter} consisting of a ViT encoder followed by a lightweight transformer-based decoder for dense prediction. 
In our setting, we use DeiT-B \cite{deit} as the encoder and keep the decoder architecture unchanged across experiments.
Compression is applied only to the encoder: token-compression methods reduce the encoder token sequence, while NViT reduces encoder architectural capacity.
For token-compression methods, the compressed token sequence is passed to the decoder and then restored to the original spatial grid for dense prediction, following prior ViT-segmentation token-compression pipelines \cite{algm,cts}.
For pruning-based models, the full spatial token grid is preserved throughout the encoder and decoder.

\paragraph{\textbf{Evaluated methods.}} 
We choose representative methods from both families:
\textit{(i) Token compression:} ToMe as a general-purpose method and ALGM, CTS as segmentation-specific methods.
\textit{(ii) Structural pruning:} NViT, which prunes architectural redundancy.
\textit{(iii) Prune-then-merge (PtM):} a two-stage combination that applies ToMe on top of an NViT-pruned encoder.
In particular, we first select a moderately pruned NViT model as the backbone, and then sweep a small set of ToMe merging rates to match the target FLOPs budget.

\paragraph{\textbf{Training and compression protocols.}} 
For every configuration, we finetune using the same training recipe~\cite{algm}, which follows the Segmenter pipeline~\cite{segmenter}, and report the best validation mIoU.
All models are initialized from ImageNet-pretrained DeiT-B or NViT checkpoints and finetuned on the target segmentation dataset.
We use SGD with momentum $0.9$, a polynomial learning-rate schedule with power $0.9$, drop-path rate $0.1$, without weight decay or learning-rate warmup.
Following the ALGM-Segmenter configuration~\cite{algm}, we train ADE20K models for $64$ epochs with batch size $8$, learning rate $10^{-3}$, and crop size $512{\times}512$.
For Cityscapes, we train for $216$ epochs with batch size $8$, learning rate $10^{-2}$, and crop size $768{\times}768$.

For token-compression methods, we follow their original compression placement: ALGM applies token merging at encoder layers $1$ and $5$, CTS performs token merging at the first encoder layer, and ToMe applies token merging across all encoder layers.
The compression is enabled during finetuning and evaluation, rather than being applied only as a post-hoc inference adaptation.
For each method, the threshold or merging rate is swept to obtain different FLOPs budgets. The ADE20K token-compression settings for the mild, moderate, and aggressive regimes are ALGM $T=\{0.94,0.75,0.65\}$, CTS $R=\{0.3,0.6,0.7\}$, and ToMe $R=\{40,95,140\}$.
The corresponding Cityscapes settings are ALGM $T=\{0.94,0.88,0.70\}$, CTS $R=\{0.4,0.5,0.65\}$, and ToMe $R=\{145,190,280\}$.

For structural pruning, we follow the NViT global-pruning protocol~\cite{nvit}. 
NViT controls pruning strength using a latency target $p$, defined relative to the original DeiT-B model.
We use the official NViT-B/H/S configurations when they match our target compression regimes, and additionally sweep $p$ over a small set of latency targets, including $p \in \{0.80,0.70,0.25\}$, to obtain FLOPs- or FPS-matched operating points for each dataset.
Unless a specific official configuration is named, we refer to these pruned models as NViT.

\paragraph{\textbf{Efficiency metrics and evaluation.}} 
Token compression changes sequence length, whereas pruning changes architectural width/structure, so their native compression ratios are not directly comparable.
We therefore use FLOPs as a common efficiency measure: for each compression technique, we measure multiple settings and compare models at similar FLOPs reductions relative to the uncompressed baseline.
We report mean Intersection-over-Union (mIoU) on clean validation images ($\text{mIoU}_{\text{clean}}$) and on corrupted validation images ($\text{mIoU}_{\text{noise}}$). 
Throughout this work, results are presented as $\text{mIoU}_{\text{clean}}/\text{mIoU}_{\text{noise}}$.
As a complementary deployment-oriented check, we report inference throughput in \cref{sec:fps}.
FPS is measured on a single NVIDIA RTX A6000 GPU with batch size $32$ and mixed precision enabled.

\subsection{Dataset and corruption protocol}
\label{sec:benchmark_dataset}
We evaluate on ADE20K and Cityscapes, together with their corrupted variants ADE20K-C and Cityscapes-C. 
For robustness evaluation, we follow the standard corruption protocol used in prior segmentation robustness work~\cite{data_c}: each validation image is transformed by a fixed set of corruption operators at multiple severity levels, and performance is evaluated under each corruption setting. 
We consider 16 corruption types grouped into four categories (Noise, Blur, Digital, Weather), each evaluated at 5 severity levels, yielding 80 corrupted variants per image. 
We report corrupted-input robustness by aggregating mIoU across corruption types and severity levels.

\section{Benchmark Results and Insights}
\label{sec:benchmark_results}
\begin{figure}[!t]
\centering
\includegraphics[width=0.49\linewidth]{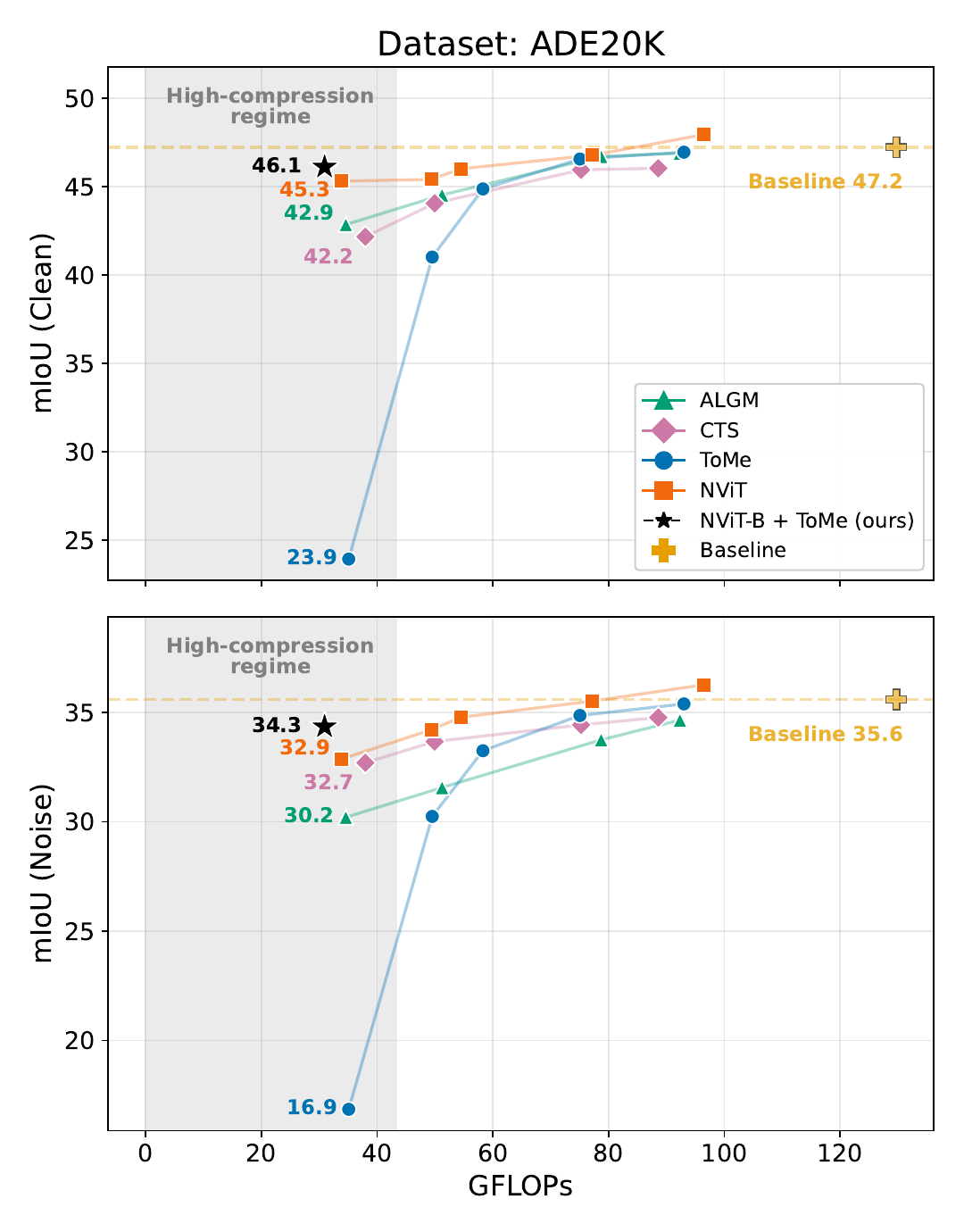}
\hfill
\includegraphics[width=0.49\linewidth]{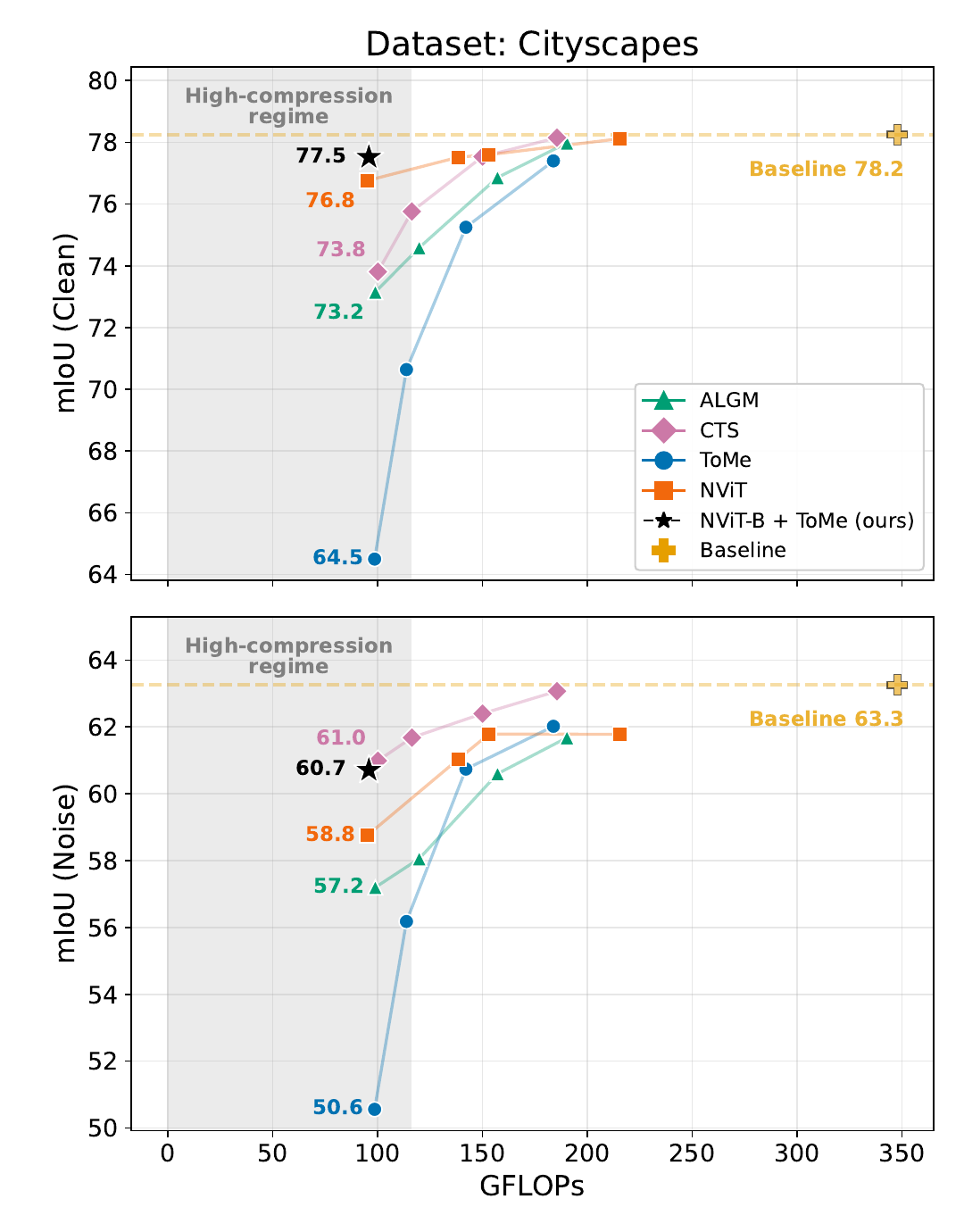}
\caption{Accuracy-compute trade-offs on ADE20K (left) and Cityscapes (right).
We compare structural pruning (NViT), token compression methods (ToMe, ALGM, CTS),
and the stack pipeline (NViT + ToMe). The top row evaluates
clean accuracy
$\text{mIoU}_{\text{clean}}$,
while the bottom row evaluates
robustness under common corruptions
$\text{mIoU}_{\text{noise}}$.
In both datasets, mild token compression can preserve performance, but aggressive token compression leads to sharp degradation; structural pruning shows a smoother degradation trend as compute decreases.}
\label{fig:tradeoff_curves}
\end{figure}
This section analyzes benchmark results under FLOPs-matched comparisons and highlights consistent patterns across compression levels.
Rather than emphasizing isolated operating points, we focus on how performance evolves as compute decreases.
Findings 1 and 2 characterize token compression across mild and aggressive regimes, while Finding 3 compares this behavior with structural pruning.
Together, these results show that token redundancy can be reduced with limited accuracy loss at mild compression, but aggressive token reduction can sharply degrade both clean performance and corruption robustness.
Structural pruning, by contrast, reduces architectural capacity while preserving the spatial token grid, leading to a more gradual degradation trend at high compression.

\subsection{Compression-regime behavior under matched FLOPs}
\label{sec:result_findings}
\cref{fig:tradeoff_curves} visualizes the overall accuracy-compute trade-offs of token compression, structural pruning, and their combination across different compute budgets on ADE20K and Cityscapes. 
Each curve shows how segmentation performance evolves as FLOPs are reduced, while \cref{tab:benchmark_main} provides detailed numeric results for each method at different compression levels.
From these results, we observe three consistent patterns across the two datasets.

\begin{table}[!t]
\centering
\resizebox{\textwidth}{!}{%
\begin{tabular}{lccc|ccc}
\toprule
& \multicolumn{3}{c}{ADE20K} & \multicolumn{3}{c}{Cityscapes} \\
\cmidrule(lr){2-4}\cmidrule(lr){5-7}
Method & GFLOPs$\downarrow$ & $\text{mIoU}_{\text{clean}}\uparrow$ & $\text{mIoU}_{\text{noise}}\uparrow$ & GFLOPs$\downarrow$ & $\text{mIoU}_{\text{clean}}\uparrow$ & $\text{mIoU}_{\text{noise}}\uparrow$ \\
\midrule
Baseline & $129.70_{\,\times 1.0}$ & 47.22 & 35.58 & $347.77_{\,\times 1.0}$ & 78.25 & 63.26 \\
\midrule
\multicolumn{7}{@{}l}{\emph{Mild compression}} \\
ALGM & $92.31_{\,\times 1.4}$ & 46.88 & 34.64 & $190.27_{\,\times 1.8}$ & 77.98 & 61.68 \\
CTS & $88.57_{\,\times 1.5}$ & 46.03 & 34.76 & $185.52_{\,\times 1.9}$ & \textbf{78.15} & \textbf{63.07} \\
ToMe & $93.00_{\,\times 1.4}$ & \underline{46.94} & \underline{35.38} & $183.83_{\,\times 1.9}$ & 77.40 & \underline{62.02} \\
NViT & $96.40_{\,\times 1.4}$ & \textbf{47.93} & \textbf{36.25} & $215.61_{\,\times 1.6}$ & \underline{78.12} & 61.78 \\
\midrule
\multicolumn{7}{@{}l}{\emph{Moderate compression}} \\
ALGM & $51.21_{\,\times 2.5}$ & 44.52 & 31.56 & $157.19_{\,\times 2.2}$ & 76.86 & 60.60 \\
CTS  & $49.93_{\,\times 2.6}$ & 44.05 & 33.66 & $150.03_{\,\times 2.3}$ & \underline{77.54} & \textbf{62.40} \\
ToMe & $49.51_{\,\times 2.6}$ & 41.01 & 30.24 & $142.11_{\,\times 2.5}$ & 75.25 & 60.74 \\
NViT-H & $49.37_{\,\times 2.6}$ & \underline{45.40} & \underline{34.20} & $138.55_{\,\times 2.5}$ & 77.52 & \underline{61.04} \\
NViT+ToMe \emph{(PtM)} & $49.02_{\,\times 2.7}$ & \textbf{46.79} & \textbf{34.83} & $141.46_{\,\times 2.5}$ & \textbf{77.58} & 60.69 \\
\midrule
\multicolumn{7}{@{}l}{\emph{Aggressive compression}} \\
ALGM & $34.58_{\,\times 3.8}$ & 42.85 & 30.20 & $98.94_{\,\times 3.5}$ & 73.15 & 57.20 \\
CTS  & $37.95_{\,\times 3.4}$ & 42.16 & 32.69 & $100.18_{\,\times 3.5}$ & 73.81 & \textbf{60.99} \\
ToMe & $35.08_{\,\times 3.7}$ & 23.94 & 16.86 & $98.60_{\,\times 3.5}$ & 64.50 & 50.56 \\
NViT-S & $33.82_{\,\times 3.8}$ & \underline{45.30} & \underline{32.86} & $95.05_{\,\times 3.7}$ & \underline{76.76} & 58.75 \\
NViT-B+ToMe \emph{(PtM)} & $34.84_{\,\times 3.7}$ & \textbf{45.71} & \textbf{34.33} & $95.87_{\,\times 3.6}$ & \textbf{77.52} & \underline{60.71} \\
\bottomrule
\end{tabular}}
\caption{Full benchmark under matched FLOPs on ADE20K and Cityscapes at three compression levels. We compare segmentation-aware token compression (ALGM, CTS), general token compression (ToMe), structural pruning (NViT), and the proposed prune-then-merge (PtM). GFLOPs subscripts give the compute-reduction factor. Best $\text{mIoU}_{\text{clean}}/\text{mIoU}_{\text{noise}}$ per dataset and level in \textbf{bold}, second best \underline{underlined}.
For NViT rows without an official B/H/S suffix, the latency target $p$ is swept per dataset to match the corresponding compute regime.
}
\label{tab:benchmark_main}
\end{table}

\paragraph{\textbf{Finding 1: Mild token compression can preserve near-baseline performance.}}
\cref{tab:benchmark_main} shows \emph{mild} token-compression performance at low compression levels. 
Across both ADE20K and Cityscapes, several token-compression configurations achieve substantial compute reductions (about $1.4\times$--$1.9\times$) while preserving near-baseline performance on both clean and corrupted inputs.
This effect is most consistent for segmentation-aware methods (CTS, ALGM), which are designed to maintain spatial information during token reduction. 
For instance, on ADE20K, CTS reduces compute from $129.70$ to $88.57$ GFLOPs ($1.5\times$), with only a small drop in $\text{mIoU}_{\text{clean}}$/\allowbreak$\text{mIoU}_{\text{noise}}$ from $47.22/35.58$ to $46.03/34.76$. 
On Cityscapes, CTS achieves a $1.9\times$ compute reduction (347.77 to 185.52 GFLOPs) with relatively similar performance ($78.15/63.07$ vs.\ $78.25/63.26$). 
Overall, these results indicate that ViT-based segmenters contain substantial token redundancy that can be reduced with limited accuracy loss at mild compression when spatial structure is preserved.

\paragraph{\textbf{Finding 2: At high compression, token compression degrades sharply.}}
At high-compression settings ($3.4\times$-$3.8\times$ FLOPs reduction), token compression drops substantially in both clean performance and corruption robustness.
As shown by the steep drops in \cref{fig:tradeoff_curves}, this degradation becomes much sharper than in the mild-compression regime.
According to \cref{tab:benchmark_main}, the degradation is particularly severe for general token merging.
ToMe collapses on ADE20K from the baseline $47.22/35.58$ to $23.94/16.86$ at $35.08$ GFLOPs ($3.7\times$ reduction), suggesting substantial loss of spatial information when an excessive number of tokens are merged across encoder layers.
On Cityscapes, ToMe also shows a large performance drop to $64.50/50.56$.
CTS and ALGM decrease more gracefully but still exhibit noticeable performance loss under extreme compression. 
For instance, CTS maintains comparatively stronger performance ($73.81/60.99$ on Cityscapes at $100.18$ GFLOPs), yet still falls below the baseline by a clear gap.

Overall, these results suggest that aggressive token reduction can discard or distort critical spatial information required for dense prediction, leading to significant degradation in both accuracy and robustness.
This behavior contrasts with structural pruning (Finding~3), which reduces model capacity more gradually and therefore exhibits a smoother degradation pattern. We provide further analysis in \cref{sec:further_analysis}.

\paragraph{\textbf{Finding 3: Model pruning is more stable at high compression.}}
\cref{fig:tradeoff_curves} shows that structural pruning exhibits a much smoother degradation pattern as compute decreases.
Unlike token compression, which can drop sharply when many tokens are merged, NViT reduces model capacity while preserving the full spatial token grid. 
Empirically, performance decreases steadily rather than collapsing.
On ADE20K, the baseline achieves $47.22/35.58$ at $129.70$ GFLOPs,
NViT achieves $47.93/36.25$, $45.40/34.20$, and $45.30/32.86$ at $96.40$ ($1.4\times$), $49.37$ ($2.6\times$), and $33.82$ ($3.8\times$) GFLOPs  respectively.
On Cityscapes, the aggressive NViT-S still achieves $76.76/58.75$ at $95.05$ GFLOPs ($3.7\times$), compared with the baseline $78.25/63.26$.
Overall, these results suggest that structural pruning provides a more stable high-compression path than aggressive token reduction.

\subsection{Prune-then-merge as a practical high-compression strategy}
\label{sec:result_methods}
The observations from Findings 1-3 suggest that token compression and structural pruning degrade differently under high compression. 
Token compression reduces token redundancy and becomes fragile when compression is aggressive, whereas pruning reduces model capacity more gradually while preserving the spatial token grid.
This motivates a strategy: apply moderate structural pruning to obtain a compact backbone, then apply moderate token merging to further reduce computation.
We therefore evaluate a \emph{prune-then-merge (PtM)} pipeline, where token merging (ToMe) is applied on top of a pruned backbone (NViT).
As shown in \cref{tab:benchmark_main}, this combined configuration achieves a strong accuracy-robustness trade-off at similar FLOPs in the high-compression regime.
At high compression, PtM achieves $45.71/34.33$ mIoU at $34.84$ GFLOPs on ADE20K and $77.52/60.71$ at $95.87$ GFLOPs on Cityscapes, corresponding to $3.7\times$ and $3.6\times$ FLOPs reduction, respectively.
It outperforms both token compression and structural pruning on ADE20K, and gives the best clean mIoU with competitive robustness on Cityscapes.
These results suggest that pruning and token compression are complementary under high compression.
Pruning first reduces model capacity while retaining the spatial layout required for dense prediction, after which moderate token merging can further reduce computation without the sharp information loss observed under aggressive token compression alone.
When applied sequentially, this pipeline provides a practical recipe for achieving strong accuracy-robustness trade-offs under high compression.

\subsection{Why does aggressive token compression break?}
\label{sec:further_analysis}
\begin{figure}[t]
  \centering
  \includegraphics[width=\linewidth]{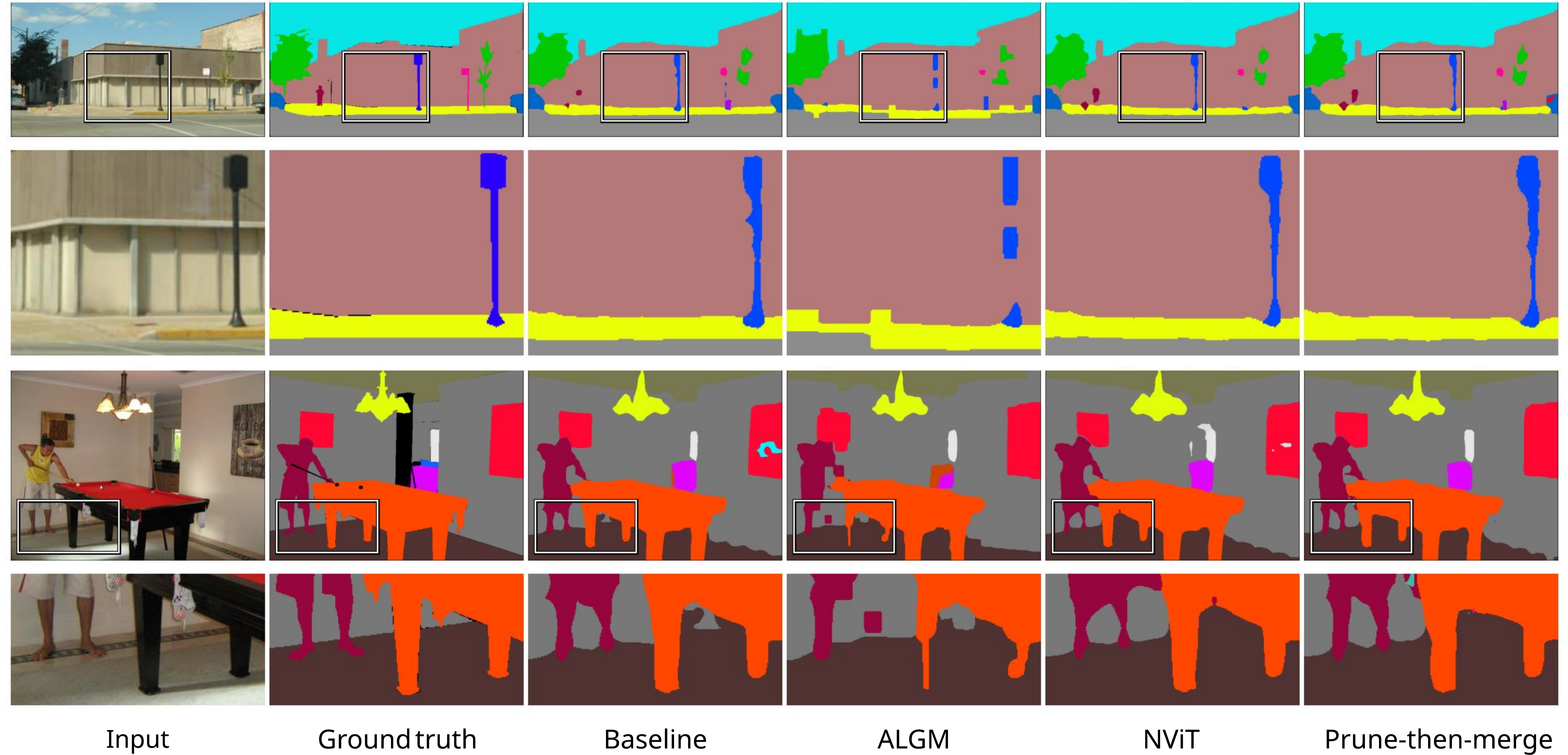}
  \caption{Qualitative comparison under aggressive compression on ADE20K.
  We show two representative examples with zoomed regions highlighting fine structures and local boundaries.
  Compared with NViT and prune-then-merge, ALGM produces less spatially coherent predictions in these regions.}
  \label{fig:visualization_aggressive}
\end{figure}
\paragraph{Effective-rank analysis.}
\begin{wrapfigure}{r}{0.48\textwidth}
    \vspace{-\intextsep}
    \centering
    \includegraphics[width=0.46\textwidth]{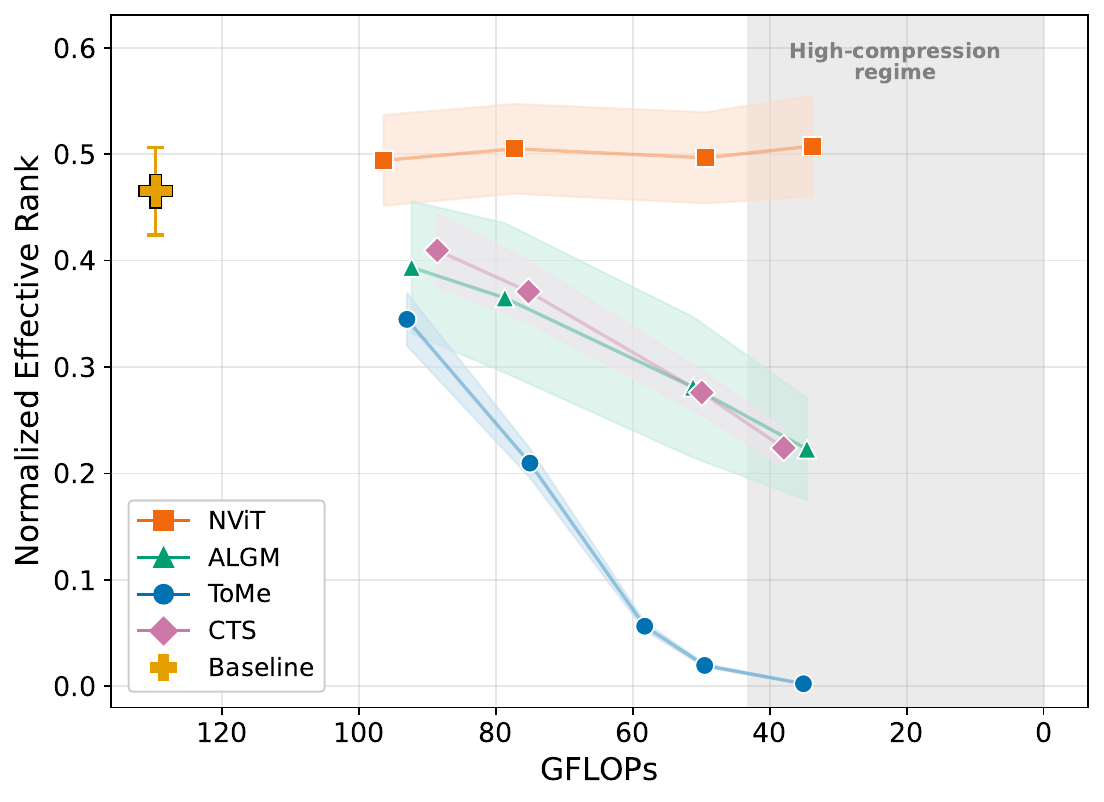}
    \caption{Normalized effective rank on ADE20K-val (markers and bands show mean $\pm$ std over images).}
    \label{fig:effective_rank}
    \vspace{-25pt}
\end{wrapfigure}
We further analyze feature diversity using entropy-based effective rank~\cite{roy2007effective} as a spectral diagnostic of representation dimensionality and collapse in dense prediction~\cite{chenlinwei2025ICCV}. 
For each image, we reconstruct the encoder features on the original token grid and denote the resulting feature matrix as \(X \in \mathbb{R}^{N \times C}\), where \(N\) is the number of tokens and \(C\) is the feature dimension.
Following~\cite{roy2007effective}, we compute \(\mathrm{eRank}(X)\) and report \(\mathrm{eRank}(X)/\min(N,C)\), which normalizes by the maximum attainable rank, and measures the fraction of available rank used by the features.
Lower values indicate that the features are concentrated along fewer dominant directions, suggesting stronger dimensional collapse~\cite{zhuo2023towards,huang2024ldreg}.

\cref{fig:effective_rank} shows that token compression methods exhibit a clear decline in normalized effective rank as compression increases. 
The drop is most severe for ToMe, indicating that aggressive merging concentrates the reconstructed encoder representation into fewer dominant feature directions. 
Segmentation-aware methods such as ALGM and CTS show a more gradual decline, but still lose feature diversity at higher compression. 
In contrast, NViT maintains a stable normalized effective rank across compression levels, suggesting that structural pruning better preserves feature diversity in this setting by retaining the spatial token grid.

\paragraph{Qualitative results.}
\cref{fig:visualization_aggressive} provides qualitative examples that illustrate the high-compression behavior observed in our quantitative results.
Under aggressive token compression, ALGM merges local tokens more aggressively, which results in less coherent predictions around thin structures and local boundaries.
In the first example, ALGM fragments the pole-like object and introduces class confusion across the sidewalk-road boundary, with sidewalk and road labels bleeding into each other.
In the second example, similar artifacts appear around the table legs and the boundaries between the person, table, and floor.
NViT and prune-then-merge maintain more coherent spatial predictions in these regions.

\subsection{FPS-matched validation}
\label{sec:fps}
\begin{table}[!t]
\centering
\footnotesize
\setlength{\tabcolsep}{1.5pt}
\begin{tabular}{lccccc}
\toprule
Method & $r_{\mathrm{FLOPs}}\uparrow$ & FPS$\uparrow$ & Speedup$\uparrow$ & $\text{mIoU}_{\text{clean}}\uparrow$ & $\text{mIoU}_{\text{noise}}\uparrow$ \\
\midrule
Baseline & $1.00\times$ & 23.69 & $1.00\times$ & 47.22 & 35.58 \\
\midrule
\multicolumn{6}{@{}l}{\emph{Mild speedup ($\sim$1.4$\times$)}} \\
ALGM & $1.41\times$ & 33.45 & $1.41\times$ & 46.88 & 34.64 \\
CTS  & $1.46\times$ & 33.69 & $1.42\times$ & 46.03 & 34.76 \\
ToMe & $1.39\times$ & 32.87 & $1.39\times$ & \textbf{46.94} & 35.38 \\
NViT & $1.68\times$ & 32.22 & $1.36\times$ & 46.77 & \textbf{35.51} \\
\midrule
\multicolumn{6}{@{}l}{\emph{Moderate speedup ($\sim$2.5$\times$)}} \\
ALGM  & $2.53\times$ & 60.01 & $2.53\times$ & 44.52 & 31.56 \\
CTS   & $2.60\times$ & 58.44 & $2.47\times$ & 44.05 & \textbf{33.66} \\
ToMe  & $2.62\times$ & 61.50 & $2.60\times$ & 41.01 & 30.24 \\
NViT-S          & $3.84\times$ & 57.54 & $2.43\times$ & \textbf{45.30} & 32.86 \\
\midrule
\multicolumn{6}{@{}l}{\emph{Aggressive speedup ($\sim$3.5$\times$)}} \\
ALGM  & $3.75\times$ & 78.90 & $3.33\times$ & 42.85 & 30.20 \\
CTS   & $3.42\times$ & 72.21 & $3.05\times$ & 42.16 & 32.69 \\
ToMe  & $3.70\times$ & 84.96 & $3.59\times$ & 23.94 & 16.86 \\
NViT  & $7.61\times$ & 87.34 & $3.69\times$ & 41.01 & 26.85 \\
NViT-B+ToMe \emph{(PtM)} & $4.75\times$ & 88.10 & $3.72\times$ & \textbf{45.17} & \textbf{33.92} \\
\bottomrule
\end{tabular}
\caption{Wall-clock comparison on ADE20K grouped by measured FPS speedup among segmentation-aware token compression (ALGM, CTS), general token compression (ToMe), structural pruning (NViT), and prune-then-merge (PtM). 
The FLOPs reduction ratio $r_{\mathrm{FLOPs}}$ and FPS speedup are reported relative to the uncompressed baseline. 
\textbf{Bold} indicates the best clean and corruption mIoU among compression methods within each speed group.
}
\label{tab:fps}
\end{table}


We further evaluate inference speed (FPS).
\cref{tab:fps} compares representative ADE20K configurations at similar wall-clock speedups.
The results show a similar regime-level trend. At mild-to-moderate speedups, ALGM, CTS, and NViT remain effective, while general token merging (ToMe) already degrades more noticeably at moderate speedups. 
At the aggressive speedup level, this gap becomes larger: ToMe drops sharply to $23.94/16.86$ mIoU at around $3.6\times$ FPS speedup, whereas ALGM and CTS token compression methods remain stronger but still lose accuracy compared to their moderate-speedup configurations. 
In contrast, prune-then-merge achieves the strongest clean/corruption trade-off among the tested high-speed configurations. At around $3.7\times$ FPS speedup, PtM (NViT-B+ToMe) maintains $45.17/33.92$ mIoU, close to the uncompressed baseline of $47.22/35.58$ and ahead of other single-axis compression methods.


We note that, 
under this segmentation pipeline, pruning-only models require much larger FLOPs reductions to achieve similar FPS speedups.
NViT-S reduces FLOPs by $3.84\times$ but achieves only a $2.43\times$ FPS speedup.
To compare pruning in the aggressive wall-clock regime, we therefore include a more aggressively pruned NViT variant, obtained with a smaller latency target.
This variant reaches a $3.69\times$ FPS speedup, but its clean/corrupted mIoU drops to $41.01/26.85$.
We hypothesize that this mismatch happens because token compression and structural pruning affect runtime computation differently. 
Following prior ViT-segmentation token-compression pipelines~\cite{algm,cts}, token compression is applied in the encoder, and the reduced token set is passed to the unchanged decoder before full spatial predictions are reconstructed. 
Pruning-only models, in contrast, reduce architectural capacity while preserving the full spatial token grid throughout.
More generally, FLOPs reduction does not necessarily yield proportional latency speedup, as wall-clock performance also depends on memory access, tensor shapes, and kernel efficiency \cite{ma2018shufflenet,HALP}.


\section{Conclusion}
Token compression has emerged as a popular approach for improving the efficiency of Vision Transformers (ViTs) while maintaining strong performance, including for semantic segmentation. 
However, existing evaluations often emphasize low-to-moderate compression, leaving behavior under high compression, where larger compute reduction may be required for deployment, less systematically analyzed.
In this work, we presented a unified benchmark of representative token compression and structural pruning methods for ViT-based semantic segmentation across a wide range of FLOPs budgets on both clean inputs and common-corruption variants. 
Our benchmark reveals a clear regime-dependent trend: token compression can preserve performance under mild reduction, but degrades sharply under aggressive compression. 
In contrast, structural pruning exhibits a smoother degradation curve and is typically more stable at high compression.
Motivated by these findings, we studied a practical \emph{prune-then-merge} pipeline that applies moderate token merging on top of a moderately pruned backbone.
Under comparable compute budgets at high compression, this strategy achieves a stronger accuracy-robustness trade-off than applying either token compression or structural pruning alone.

Future work includes extending the benchmark to additional recognition tasks, such as classification and detection, and to broader backbone families beyond flat-token ViTs, such as hierarchical transformers commonly used for high-resolution vision. 
In addition, we aim to study more principled ways to co-design pruning and token compression for improved robustness at aggressive compression levels.


\section*{Acknowledgements}
This research is supported by Temasek Laboratories, Singapore University of Technology and Design;
the National Research Foundation, Singapore under its AI Singapore Programmes (AISG Award No.: AISG2-TC-2022-007); 
the Agency for Science, Technology and Research (A*STAR) under its MTC Programmatic Funds (Grant No. M23L7b0021);
the National Research Foundation, Singapore and Infocomm Media Development Authority under its Trust Tech Funding Initiative. Any opinions, findings and conclusions or recommendations expressed in this material are those of the author(s) and do not reflect the views of National Research Foundation, Singapore and Infocomm Media Development Authority.

%
%
\bibliographystyle{splncs04}
\bibliography{main}
\end{document}